\documentclass[runningheads]{llncs}
\usepackage{graphicx}
\usepackage{amssymb}
\usepackage{amsmath}
\usepackage{threeparttable}
\usepackage[marginal]{footmisc}
\usepackage{multirow}
\usepackage{comment}
\usepackage{bbding}
\usepackage[colorlinks, anchorcolor=blue, linkcolor=red, citecolor=blue]{hyperref} 
\begin{document}
%
\title{Learning Motion Flows for Semi-supervised Instrument Segmentation from \\ Robotic Surgical Video}
\titlerunning{Learning Motion Flows for Semi-supervised Instrument Segmentation}
%

\author{Zixu Zhao\inst{1}\and
	Yueming Jin  \inst{1}\Envelope \and
	Xiaojie Gao\inst{1} \and
	Qi Dou\inst{1,2} \and
	Pheng-Ann Heng\inst{1,3}}
\authorrunning{Z. Zhao et al.}
\institute{Department of Computer Science and Engineering,\\
	The Chinese University of Hong Kong, Hong Kong, China \\ \email{ \{zxzhao, ymjin\}@cse.cuhk.edu.hk} \\\and
	Shun Hing Institute of Advanced Engineering, CUHK, Hong Kong, China\\\and
	T Stone Robotics Institute, CUHK, Hong Kong, China \\
	}

\maketitle              

\begin{abstract}
Performing low hertz labeling for surgical videos at intervals can  greatly releases the burden of surgeons.
In this paper, we study the semi-supervised instrument segmentation from robotic surgical videos with sparse annotations.
Unlike most previous methods using unlabeled frames individually, we propose a dual motion based method to wisely learn motion flows for segmentation enhancement by leveraging temporal dynamics.
We firstly design a flow predictor to derive the motion for jointly propagating the frame-label pairs given the current labeled frame. 
Considering the fast instrument motion, we further introduce a flow compensator to estimate intermediate motion within continuous frames, with a novel cycle learning strategy. 
By exploiting generated data pairs, our framework can recover and even enhance temporal consistency of training sequences to benefit segmentation.
We validate our framework with binary, part, and type tasks on 2017 MICCAI EndoVis Robotic Instrument Segmentation Challenge dataset. 
Results show that our method outperforms the state-of-the-art semi-supervised methods by a large margin, and even exceeds fully supervised training on two tasks\footnote[4]{Our code is available at \url{https://github.com/zxzhaoeric/Semi-InstruSeg/} }.

\end{abstract}

\section{Introduction}

By providing the context perceptive assistance, semantic segmentation of surgical instrument can greatly benefit robot-assisted minimally invasive surgery towards superior surgeon performance.
Automatic instrument segmentation also serves as a crucial cornerstone for  more downstream capabilities such as tool pose estimation~\cite{kurmann2017simultaneous}, tracking and control~\cite{du2019patch}. 
Recently, convolutional neural network (CNN) has demonstrated new state-of-the-arts on surgical instrument segmentation thanks to the effective data-driven learning ~\cite{garcia2017toolnet,milletari2018cfcm,shvets2018automatic,jin2019incorporating}.
However, these methods highly rely on abundant labeled frames to achieve the full potential.
It is expensive and laborious especially for high frequency robotic surgical videos, entailing the frame-by-frame pixel-wise annotation by experienced experts.

Some studies tend to utilize extra signals to generate parsing masks, such as robot kinematic model~\cite{da2019self,qin2019surgical}, weak annotations of object stripe and skeleton~\cite{fuentes2019easylabels}, and simulated surgical scene~\cite{pfeiffer2019generating}. 
However, additional efforts are still required for other signal access or creation.
Considerable effort has been devoted to utilizing the large-scale unlabeled data to improve segmentation performance for medical image analysis~\cite{zhang2017deep,bai2017semi,yu2019uncertainty}. 
For example, Bai et al.~\cite{bai2017semi} propose a self-training strategy for cardiac segmentation, where the supervised loss and semi-supervised loss are alternatively updated.
Yu et al.~\cite{yu2019uncertainty} raise an uncertainty-aware mean-teacher framework for 3D left atrium segmentation by learning the  reliable knowledge from unlabeled data.
In contrast, works focusing on the effective usage of unlabeled surgical video frames are limited. 
The standard mean teacher framework has recently been applied to the semi-supervised liver segmentation by computing the consistency loss of  laparoscopic images~\cite{fu2019more}. 
Ross et al.~\cite{ross2018exploiting} exploit a self-supervised strategy by using GAN-based re-colorization on individual unlabeled endoscopic frames for model pretraining.

Unfortunately, these semi-supervised methods propose to capture the information based on separate unlabeled video frames, failing to leverage the inherent temporal property of surgical sequences.
Given 50\% labeled frames with a labeling interval of 2, a recent approach~\cite{jin2019incorporating} indicates that utilizing temporal  consistency of surgical videos benefit semi-supervised segmentation.
Optical flows are used to transfer predictions of unlabeled frames to adjacent position whose labels are borrowed to calculate semi-supervised loss.
Yet this method heavily depends on accurate optical flow estimation and fails to provide trustworthy semi-supervision in model with some erroneous transformations.

In this paper, we propose a dual motion based semi-supervised framework for instrument segmentation by leveraging the self-contained sequential cues in surgical videos. 
Given sparsely annotated sequences, our core idea is to derive the motion flows for annotation and frame transformation that recover the temporal structure of raw videos to boost semi-supervised segmentation.
Specifically, we firstly design a flow predictor to learn the motion  between two frames with a video reconstruction task. 
We propose a joint propagation strategy to generate  frame-label pairs with learned flows, alleviating the misalignment of pairing propagated labels with raw frames.
Next, we design a flow compensator with a frame interpolation task to learn the intermediate motion flows. 
A novel unsupervised cycle learning strategy is proposed to optimize models by minimizing the discrepancy between forward predicted frames and backward cycle reconstructions.
The derived motion flows further propagate intermediate frame-label pairs as the augmented data to  enhance the sequential consistency.
Rearranging the training sequence by replacing unlabeled raw frames with generated data pairs, our framework can greatly  benefit segmentation performance. 
We extensively evaluate the method on surgical instrument binary, part, and type segmentation tasks on 2017 MICCAI EndoVis Challenge dataset.  Our method consistently outperforms state-of-the-art semi-supervised segmentation methods by a large margin, as well as exceeding the fully supervised training on two tasks.

\section{Method}
\begin{figure}[th]
	\centering
	\includegraphics[width=\textwidth]{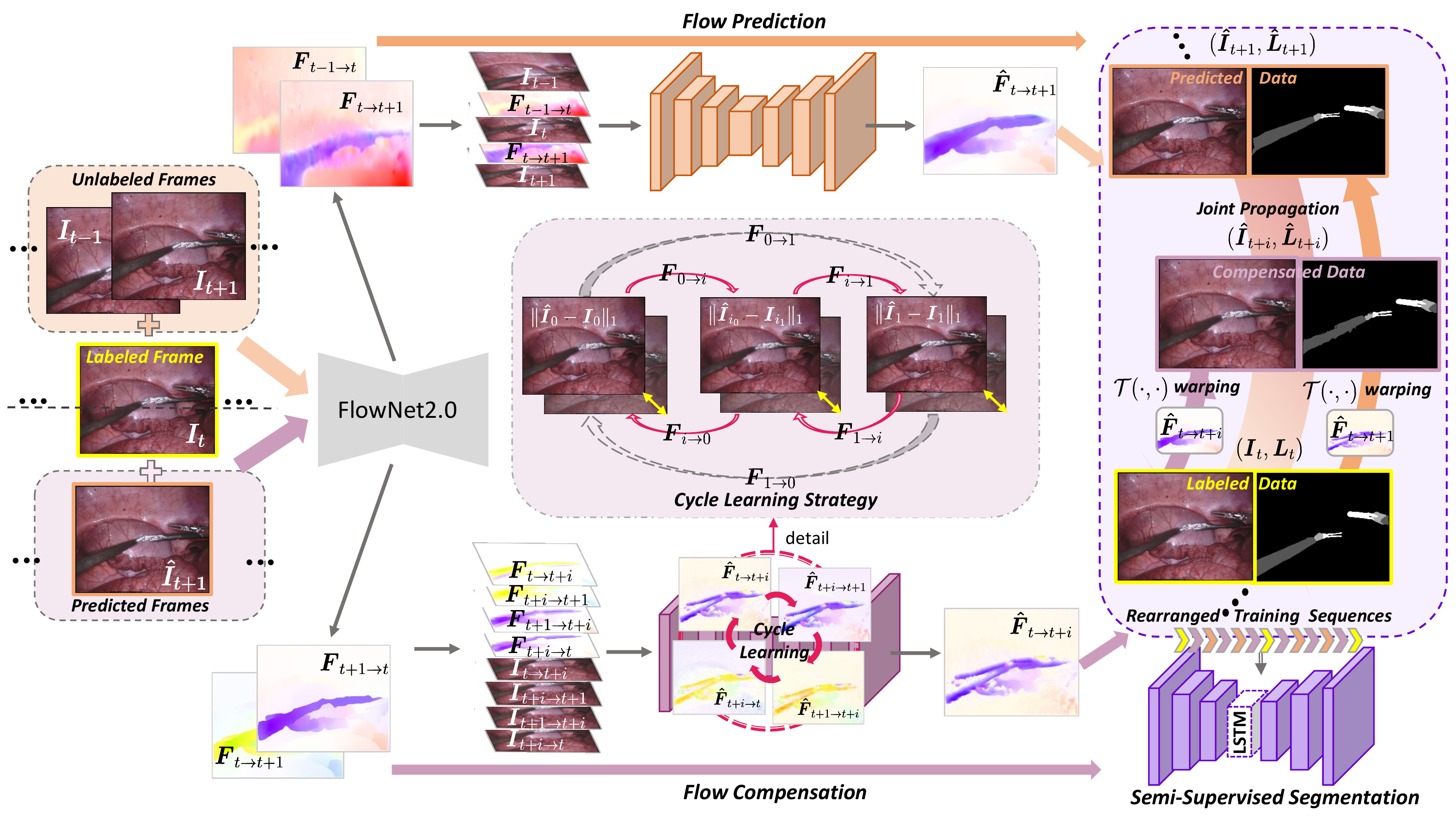}
	\caption{The illustration of the proposed framework. We learn motion flows along \textit{flow prediction} branch and \textit{flow compensation} branch successively, which are used to joint propagate the aligned data pairs for semi-supervised segmentation.  }
	\label{fig:framework}
\end{figure}

Fig.~\ref{fig:framework} illustrates our dual motion-based framework. It uses raw frames to learn dual motion flows, one for recovering original annotation distribution (top branch) and the other for compensating fast instrument  motions (bottom branch). We ultimately use learned motion flows to propagate aligned frame-label pairs as a substitute for unlabeled raw frames in video sequences for segmentation training.
\subsection{Flow Prediction for Joint Propagation}
With a video having $T$ frames as $ \boldsymbol{I} =\{I_{0}, I_{1}, ..., I_{T-1}\} $, we assume that $\boldsymbol{I}$ is labeled with intervals. 
For instance, only $ \{I_{0}, I_{5}, I_{10},...\} $ are labeled with interval 4, accounting for 20\% labeled data.
This setting is reasonable to clinical practice, as it is easier for surgeons to perform low hertz labeling.
Sharing the spirit with \cite{jin2019incorporating}, we argue that the motion hidden within the continuous raw frames can be applied to corresponding instrument masks.
Therefore, we first derive the motion flow from raw frames with a video reconstruction task, as shown in \textit{Flow  Prediction} branch in Fig.~\ref{fig:framework}.
Given the sequence $I_{t':t+1}$, we aim to estimate the motion flow $\hat{F}_{t\to t+1}$ that can translate the current frame $I_t$ to future frame $I_{t+1}$:
\begin{equation}
\label{pred}
\hat{F}_{t\to t+1} = \mathcal{G}(I_{t':t+1}, F_{t'+1:t+1}), ~~ \hat{I}_{t+1}=\mathcal{T}(I_{t}, \hat{F}_{t\to t+1}), 
\end{equation}
where $\mathcal{G} $ is a 2D CNN based flow predictor
with the input $I_{t':t}$. Optical flows $F_i$ between successive frames $I_i$ and $I_{i-1}$ are calculated by FlowNet2.0~\cite{ilg2017flownet}. 
$\mathcal{T} $ is a forward warping function which is differentiable and implemented with bilinear interpolation~\cite{zhou2016view}.
Instead of straightforwardly relying on the optical flow \cite{jin2019incorporating}, which suffers from the undefined problem for the dis-occluded pixels,
we aim to learn the motion vector $(u, v)$ as a precise indicator for annotation propagation which can effectively account for the gap.
Intuitively, the instrument mask follows the same location shift as its frame. 
For an unlabeled frame $I_{t+1}$, we can borrow the adjacent annotation $L_{t}$ and use the derived flow for its label propagation:
\begin{equation}
\hat{L}_{t+1}=\mathcal{T}(L_{t}, \hat{F}_{t\to t+1}).
\end{equation}
Directly pairing the propagated label with the original future frame ($I_{t+1}, \hat{L}_{t+1}$) for our semi-supervised segmentation may encounter the mis-alignment issue in the region whose estimated motion flows are inaccurate. Motivated by~\cite{zhu2019improving}, we introduce the concept of joint propagation into our semi-supervised setting. We pair the propagated label with predicted future frame ($\hat{I}_{t+1}, \hat{L}_{t+1}$), while leaving the original data $I_{t+1}$  merely for motion flow generation.
Such joint propagation avoids introducing the erroneous regularization towards network training.
Furthermore, we can  bi-directionally apply the derived motion flow with multiple steps, obtaining ($\hat{I}_{t-k:t+k}, \hat{L}_{t-k:t+k}$) with $k$ steps ($k=1,2, 4$ in our experiments).
The superior advantage of joint propagation can be better demonstrated when performing such multi-step propagation in a video with severely sparse annotations, as it alleviates accumulating errors within derived motion flows.\\
\textbf{Supervised Loss Functions.}  The overall loss function of flow predictor is:
\begin{equation}
	\mathcal{L}_{Pred} = \lambda_{1}\mathcal{L}_{1} +\lambda_{p}\mathcal{L}_{p} +\lambda_{s}\mathcal{L}_{s},
\end{equation}
consisting of the primary loss, i.e., L1 loss $ \mathcal{L}_{1}=\lVert\hat{I}_{t+1}-I_{t+1}\lVert_{1} $, which can capture subtle modification rather than L2 loss~\cite{reda2018sdc}; perceptual loss $\mathcal{L}_{p}$ to retain structural details in predictions, detailed definition in~\cite{johnson2016perceptual}; smooth loss $\mathcal{L}_{s}= \lVert\nabla F_{t\to t+1}\lVert_{1}$ to encourage neighboring pixels to have similar flow values~\cite{jiang2018super}.
We empirically set the weights as  $\lambda_{1} =0.7 $, $\lambda_{p} =0.2 $, and $\lambda_{s} =0.1 $ for a more robust combination in eliminating the artifacts and occlusions than~\cite{reda2018sdc}. 

\subsection{Flow Compensation via Unsupervised Cycle Learning}
The fast instrument motions between successive frames always occur even in a high frequency  surgical video.
Smoothing large motions thus improves the sequential consistency, as well as adds data variety for semi-supervised segmentation. In this scenario, we try to compensate motion flows with a frame interpolation task. However, existing interpolation approaches are not suitable for our case. Either the optical flow based methods~\cite{reda2019unsupervised,jiang2018super} rely on consistent motion frequency, or the kernel based method~\cite{niklaus2017video} contradicts the alignment prerequisite.
Hence, we propose an unsupervised flow compensator with a novel cycle learning strategy, which forces the model to learn intermediate flows by minimizing the discrepancy between  forward predicted frames and their backward cycle reconstructions.

Given two continuous frames $I_{0}$ and $I_{1}$, our ultimate goal is to learn a motion flow $ \hat{F}_{0\to i} $ with a time $i\in(0,1)$ to jointly propagate the intermediate frame-label pair ($\hat{I}_{i}$,$\hat{L}_{i}$).
In the \textit{Flow Compensation} branch, we first use the pretrained FlowNet2.0 to compute  bi-directional optical flows ($F_{0\to1}$, $F_{1\to0}$) between two frames.
We then use them to approximate the intermediate optical flows $\boldsymbol{F}$:
\begin{equation}
\label{eq:1}
\begin{aligned}
F_{i\to 0}&=-(1-i)iF_{0\to 1} +i^{2}F_{1\to 0}, ~~~~~~ F_{1\to i} &=F_{1\to 0}-F_{i\to 0},\\
F_{i\to 1}&=(1-i)^{2}F_{0\to 1} -i(1-i)F_{1\to 0}, ~ F_{0\to i}&=F_{0\to 1}-F_{i\to 1}.\\
\end{aligned}
\end{equation}
Such approximation suits well in smooth regions but poorly around  boundaries, however, it can still serve as an essential initialization for  subsequent cycle learning.
The approximated flows $\boldsymbol{F}$ are used to generate  warped frames $\boldsymbol{\hat{I}}$, including forward predicted frames $\hat{I}_{i_{0}}$, $\hat{I}_{1}$ and backward reconstructed frames $\hat{I}_{i_{1}}$, $\hat{I}_{0}$:
\begin{equation}
\begin{aligned}
\hat{I}_{i_{0}}= \mathcal{T}(I_{0}, F_{0\to i}),
\hat{I}_{1}= \mathcal{T}(\hat{I}_{i_{0}}, F_{i\to 1}),
\hat{I}_{i_{1}}=
\mathcal{T}(\hat{I}_{1}, F_{1\to i}),
\hat{I}_{0}= \mathcal{T}(\hat{I}_{i_{1}}, F_{i\to 0}). \\
\end{aligned}
\end{equation}
We then establish a flow compensator that based on a 5-stage U-Net to refine motion flows with cycle consistency.
It takes the two frames ($ I_{0} $, $ I_{1} $), four initial approximations $\boldsymbol{F}$, and four warped frames $\boldsymbol{\hat{I}}$ as input, and outputs four refined flows $ \boldsymbol{\hat{F}} $, where $\hat{F}_{0\to i}$ is applied on $I_{0}$ for joint frame-label  pair generation.
\\
\textbf{Unsupervised Cycle Loss.} 
The key idea is to learn the motion flow that can encourage models to satisfy cycle consistency in time domain.
Intuitively, we expect that the predicted $\hat{I}_{1}$ and reconstructed $\hat{I}_{0}$ are well overlapped with the original raw data $I_1$ and $I_0$. Meanwhile, two intermediate frames warped along a cycle, i.e., $\hat{I}_{i_{1}}$ and $\hat{I}_{i_{0}}$, should show the similar representations.
Keeping this in mind, we use L1 loss to primarily constrain the inconsistency of each pair:
\begin{equation}
\begin{aligned}
\mathcal{L}_{1}^c=\lambda_{0}\lVert\hat{I}_{0}-I_{0}\lVert_{1} + \lambda_{i}\lVert\hat{I}_{i_{0}}-\hat{I}_{i_{1}}\lVert_{1} +\lambda_{1}\lVert\hat{I}_{1}-I_{1}\lVert_{1}.
\end{aligned}
\end{equation}
To generate sharper predictions, we add the perceptual loss $\mathcal{L}_{p}^c$ on the three pairs (perceptual loss definition in \cite{johnson2016perceptual}).
Our overall unsupervised cycle loss is defined as $\mathcal{L}_{cycle} = \mathcal{L}_{1}^c + \lambda_{p}\mathcal{L}_{p}^c$,
where we empirically set $\lambda_{0} =1.0 $, $\lambda_{i} =0.8 $, $\lambda_{1} =2.0 $, and  $\lambda_{p} =0.01 $. 
Our cycle regularization can avoid relying on the immediate frames and learn the motion flow in a completely unsupervised way.

\subsection{Semi-supervised Segmentation}
For semi-supervised segmentation, we study the sparsely annotated video sequences $\boldsymbol{I} =\{I_{0}, I_{1}, ..., I_{T-1}\}$ with a label interval $h$.
The whole dataset consists of labeled subset $ \mathcal{D}_{L}=\{(I_{t}, L_{t})\}_{t=hn}$ with $N$ frames and unlabeled subset $\mathcal{D}_{U}=\{I_{t}\}_{t \neq hn} $ with $M = hN$ frames.
Using consecutive raw frames, our flow predictor learns motion flows with a video reconstruction task, which are used to transfer the adjacent annotations for the unlabeled data.
With the merit of joint propagation, we pair the generated labels and frames, obtaining the re-labeled set $\mathcal{D}_{R}=\{\hat{I}_{t},  \hat{L}_{t} \}_{t \neq hn} $ with $M$ frames.
Subsequently, our flow compensator learns the intermediate motion flow with an unsupervised video interpolation task. 
We can then extend the dataset by adding $\mathcal{D}_{C}=\{\tilde{I}_{t_0},  \tilde{L}_{t_0} \}_{t=1}^{T-1} $ with $N$+$M$-$1$ compensated frames with  interpolation rate as 1.
Our flow predictor and compensator are designed based on U-Net, with network details in supplementary.
We finally consider $\mathcal{D}_{L}\cup \mathcal{D}_{R} \cup \mathcal{D}_{C}$ as the training set for semi-supervised segmentation. 
For the network architecture, we basically adopt the same backbone as~\cite{jin2019incorporating}, i.e., U-Net11~\cite{ronneberger2015u} with pretrained encoders from VGG11~\cite{simonyan2014very}. 
Excitingly, different from other semi-supervised methods, our motion flow based strategy retains and even enhances the inherent sequential consistency.
Therefore, we can still exploit temporal units, such as adding convolutional long short term memory layer (ConvLSTM) at the bottleneck, to increase segmentation performance.

\section{Experiments}
\textbf{Dataset and Evaluation Metrics.} We validate our method on the public dataset of Robotic Instrument Segmentation from 2017 MICCAI EndoVis Challenge~\cite{allan20192017}. 
The video sequences with a high resolution of 1280$ \times $1024 are acquired from \emph{da Vinci Xi} surgical system during different porcine procedures. 
We conduct all three sub-tasks of this dataset, i.e., binary (2 classes), part (4 classes) and type (8 classes), with gradually fine-grained segmentation for an instrument.
For direct and fair comparison, we follow the same evaluation manner in~\cite{jin2019incorporating}, by using the released 8$ \times $225-frame videos for 4-fold cross-validation, also with the same fold splits. 
Two metrics are adopted to quantitatively evaluate our method, including mean intersection-over-union (IoU) and Dice coefficient (Dice).
\\
\textbf{Implementation Details.} The framework is implemented in Pytorch with NVIDIA Titan Xp GPUs. The parameters of  pretrained FlowNet2.0 are frozen while training the overall framework with Adam optimizer.
The learning rate is  set as $1e\!-\!3$ and divided by 10 every 150 epochs. 
We randomly crop 448$ \times $448 sub-images as the framework input. 
For training segmentation models, we follow the rules in~\cite{jin2019incorporating}. 
As for the ConvLSTM based variant, the length of input sequence is 5. The initial learning rate is set as $1e\!-\!4$ for ConvLSTM layer while $1e\!-\!5$ for other network components. All the experiments are repeated 5 times to account for the stochastic nature of DNN training.
\\
\textbf{Comparison with Other Semi-supervised Methods.} 
We implement several state-of-the-art semi-supervised segmentation methods for comparison, including ASC~\cite{niklaus2017video} (interpolating labels with adaptive separable convolution), MF-TAPNet~\cite{jin2019incorporating} (propagating labels with optical flows), self-training method~\cite{bai2017semi}, 
\begin{table}[]
	\caption{Comparison of instrument segmentation results on three tasks (mean$\pm$std). }
	\label{tab2}
	\centering
	\resizebox{120mm}{!}{
	\begin{threeparttable}
	\begin{tabular}{l|c|c|c|c|c|c|cc}
		\hline
		\multirow{2}{*}{Methods} & \multicolumn{2}{c|}{Frames used} & \multicolumn{2}{c|}{Binary segmentation} & \multicolumn{2}{c|}{Part segmention} & \multicolumn{2}{c}{Type segmentation}                      \\ \cline{2-9} 
		& Label        & Unlabel       & IoU (\%)            & Dice (\%)          & IoU (\%)          & Dice (\%)         & \multicolumn{1}{c|}{IoU (\%)}          & Dice (\%)         \\ \hline
		U-Net11   & 100\%          & 0               & 82.55$ \pm $12.51   & 89.76$ \pm $9.10   & 64.87$ \pm $14.46 & 76.08$ \pm $13.05 & \multicolumn{1}{c|}{36.83$ \pm $26.36} & 45.48$ \pm $28.16 \\
        U-Net11$ ^{\star} $               & 100\%          & 0               & 83.17$ \pm $12.01   & 90.22$ \pm $8.83   & 64.96$ \pm $14.12 & 76.57$ \pm $12.44 & \multicolumn{1}{c|}{40.31$ \pm $24.38} & 49.57$ \pm $25.39 \\ \hline
        TernausNet~\cite{shvets2018automatic}    & 100\%          & 0               & 83.60$ \pm $15.83   & 90.01$ \pm $12.50   & 65.50$ \pm $17.22 & 75.97$ \pm $16.21 & \multicolumn{1}{c|}{33.78$ \pm $19.16} & 44.95$ \pm $22.89 \\ 
        MF-TAPNet~\cite{jin2019incorporating}              & 100\%           & 0    & 87.56$ \pm $16.24   & 93.37$ \pm $12.93  & 67.92$ \pm $16.50 & 77.05$ \pm $16.17& \multicolumn{1}{c|}{36.62$ \pm $22.78} & 48.01$ \pm $25.64 \\ \hline
		ASC~\cite{niklaus2017video}       & 20\%           & 80\%            & 78.51$ \pm $13.40   & 87.17$ \pm $9.88   & 59.07$ \pm $14.76 &70.92$ \pm $13.97 & \multicolumn{1}{c|}{30.19$ \pm $17.65} & 41.70$ \pm $20.62 \\
			ASC$ ^{\star} $               & 20\%           & 80\%  & 78.33$ \pm $12.67 & 87.04$ \pm $12.85   &58.93$ \pm $14.61 &  70.76$ \pm $13.40         & \multicolumn{1}{c|}{30.60$ \pm $16.55}      & 41.88$ \pm $22.24\\ 
		Self-training~\cite{bai2017semi}      & 20\%           & 80\%   & 79.32$ \pm $12.11   & 87.62$ \pm $9.46   & 59.30$ \pm $15.70 & 71.04$ \pm $14.04 & \multicolumn{1}{c|}{31.00$ \pm $25.12} & 42.11$ \pm $24.52 \\
		Re-color~\cite{ross2018exploiting}      & 20\%           & 80\%   &79.85$ \pm $13.55&  87.78$ \pm $10.10    &  59.67$ \pm $15.14   &  71.51$\pm $15.13  & \multicolumn{1}{c|}{30.72$ \pm $25.66} &41.47$ \pm $25.30  \\
		MF-TAPNet~\cite{jin2019incorporating}              & 20\%           & 80\%            & 80.06$ \pm $13.26   & 87.96$ \pm $9.57   & 59.62$ \pm $16.01 & 71.57$ \pm $15.90 & \multicolumn{1}{c|}{31.55$ \pm $18.72} & 42.35$ \pm $22.41 \\ 
		UA-MT~\cite{yu2019uncertainty}             & 20\%           & 80\%            &   80.68$ \pm $12.63                  &  88.20$ \pm $9.61  & 60.11$ \pm $14.49  &72.18$ \pm$ 13.78  & \multicolumn{1}{c|}{32.42$ \pm $21.74}    & 43.61$ \pm $26.30  \\ 
		\textbf{Our Dual MF}                     & 20\%           & 80\%            & \textbf{83.42$ \pm $12.73}    & \textbf{90.34$ \pm $9.25 }  &  \textbf{61.77$ \pm $14.19}     &  \textbf{73.22$ \pm $13.25}  & \multicolumn{1}{c|}{\textbf{37.06$ \pm $25.03}}                  & \textbf{46.55$ \pm $27.10}\\
		
		\textbf{Our Dual MF}$ ^{\star} $                & 20\%           & 80\%            & \textbf{84.05$ \pm $13.27 }   & \textbf{91.13$ \pm $9.31 }  & \textbf{62.51$ \pm $13.32}  &\textbf{74.06$ \pm $13.08}                   & \multicolumn{1}{c|}{\textbf{43.71$ \pm $25.01}}                  & \textbf{52.80$ \pm $26.16} \\ \hline\hline
			U-Net11                    & 30\%           & 0               & 80.16$ \pm $13.69   & 88.14$ \pm $10.14  & 61.75$ \pm $14.40 & 72.44$ \pm $13.41 & \multicolumn{1}{c|}{31.96$ \pm $27.98} & 38.52$ \pm $31.02 \\
	Our Single MF                & 30\%           & 70\%            & 83.70$ \pm $12.47   & 90.46$ \pm $8.95   & 63.02$ \pm $14.80 & 74.49$ \pm $13.76 & \multicolumn{1}{c|}{39.38$ \pm $25.54} & 48.49$ \pm $26.92 \\
	Our Dual MF                & 30\%           & 70\%            & \textbf{84.12$ \pm $13.18}   &\textbf{ 90.77$ \pm $9.45 }  & \textbf{63.82$ \pm $15.63} & \textbf{74.74$ \pm $13.84} & \multicolumn{1}{c|}{\textbf{39.61$ \pm $26.45}} & \textbf{48.80$ \pm $27.67} \\
	Our Dual MF$ ^{\star} $           & 30\%           & 70\%            & \textbf{84.62$ \pm $13.54}   & \textbf{91.63$ \pm $9.13 }  & \textbf{64.89$ \pm $13.26} & \textbf{76.33$ \pm $12.61} & \multicolumn{1}{c|}{\textbf{45.83$ \pm $21.96}} & \textbf{56.11$ \pm $22.33} \\ \hline
	U-Net11                    & 20\%           & 0               & 76.75$ \pm $14.69   & 85.75$ \pm $11.36  & 58.50$ \pm $14.65 & 70.70$ \pm $13.95 & \multicolumn{1}{c|}{23.53$ \pm $24.84} & 26.74$ \pm $27.17 \\
	Our Single MF                & 20\%           & 80\%            & 83.10$ \pm $12.18   & 90.15$ \pm $8.83   & 61.20$ \pm $14.10 & 72.49$ \pm $12.94 & \multicolumn{1}{c|}{36.72$ \pm $23.62} & 46.50$ \pm $25.09 \\
    Our Dual MF                & 20\%           & 80\%            & \textbf{83.42$ \pm $12.73}   & \textbf{90.34$ \pm $9.25}   & \textbf{61.77$ \pm $14.1}9 & \textbf{73.22$ \pm $13.25} & \multicolumn{1}{c|}{\textbf{37.06$ \pm $25.03}} & \textbf{46.55$ \pm $27.19} \\
	Our Dual MF$ ^{\star} $          & 20\%           & 80\%            & \textbf{84.05$ \pm $13.27}   & \textbf{91.13$ \pm $9.31}   & \textbf{62.51$ \pm $13.32} & \textbf{74.06$ \pm $13.08} & \multicolumn{1}{c|}{\textbf{43.71$ \pm $25.01}} &\textbf{ 52.80$ \pm $26.16} \\ \hline
	U-Net11                    & 10\%           & 0               & 75.93$ \pm $15.03   & 85.09$ \pm $11.77  & 55.24$ \pm $15.27 & 67.78$ \pm $14.97 & \multicolumn{1}{c|}{15.87$ \pm $16.97} & 19.30$ \pm $19.99 \\
	Our Single MF                & 10\%           & 90\%            & 82.05$ \pm $14.35   & 89.23$ \pm $10.65  & 57.91$ \pm $14.51 & 69.28$ \pm $14.79 & \multicolumn{1}{c|}{30.24$ \pm $21.33} & 40.12$ \pm $24.21 \\
	Our Dual MF               & 10\%           & 90\%            & \textbf{82.70$ \pm $13.21}   & \textbf{89.74$ \pm $9.56 }  & \textbf{58.29$ \pm $15.60} & \textbf{69.54$ \pm $15.23} & \multicolumn{1}{c|}{\textbf{31.28$ \pm $19.53}} & \textbf{41.01$ \pm $21.91} \\
	Our Dual MF$ ^{\star} $          & 10\%           & 90\%            & \textbf{83.10$ \pm $12.45}   & \textbf{90.02$ \pm $8.80}   & \textbf{59.36$ \pm $14.38} & \textbf{70.20$ \pm $13.96} & \multicolumn{1}{c|}{\textbf{33.64$ \pm $20.19}}    & \textbf{43.20$ \pm $22.70} \\ \hline
		
	\end{tabular}
	\begin{tablenotes}
				\footnotesize
				\item Note: $ ^{\star} $ denotes that the temporal unit ConvLSTM is added at the bottleneck of the segmentation  network.
			\end{tablenotes}
		\end{threeparttable}
}
\end{table}
Re-color~\cite{ross2018exploiting} (GAN-based re-colorization for model initialization), and UA-MT~\cite{yu2019uncertainty}  (uncertainty-aware mean teacher). 
We conduct experiments under the setting of 20\% frames being labeled with annotation interval as 4.
Most above methods are difficult to gain profit from temporal units except ASC, due to the uncontinuous labeled input or network design.
We use the same network backbone (U-Net11) among these methods for fair comparison. Table~\ref{tab2} compares our segmentation results with other semi-supervised methods. 
We also report fully supervised results of U-Net11 as upper bound, as well as two benchmarks TernausNet~\cite{shvets2018automatic}, and MF-TAPNet for reference. 
Among the semi-supervised methods, UA-MT achieves slightly better performance as it draws out more reliable information from unlabeled data. 
Notably, our method consistently outperforms UA-MT across three tasks by a large margin, i.e., 2.68\% in IoU and 2.24\% in Dice on average. 
After adding the temporal unit, results of ASC degrade instead on two tasks due to the inaccurate interpolated labels.  
As our semi-supervised method can enhance sequential consistency by expanded frame-label pairs, our results can be further improved with ConvLSTM, even surpassing the fully supervised training (U-Net11$ ^{\star} $) by 0.91\% Dice on binary task and 3.23\% Dice on type task. 
\begin{figure}[!h]
	\centering
	\includegraphics[width=0.85\textwidth]{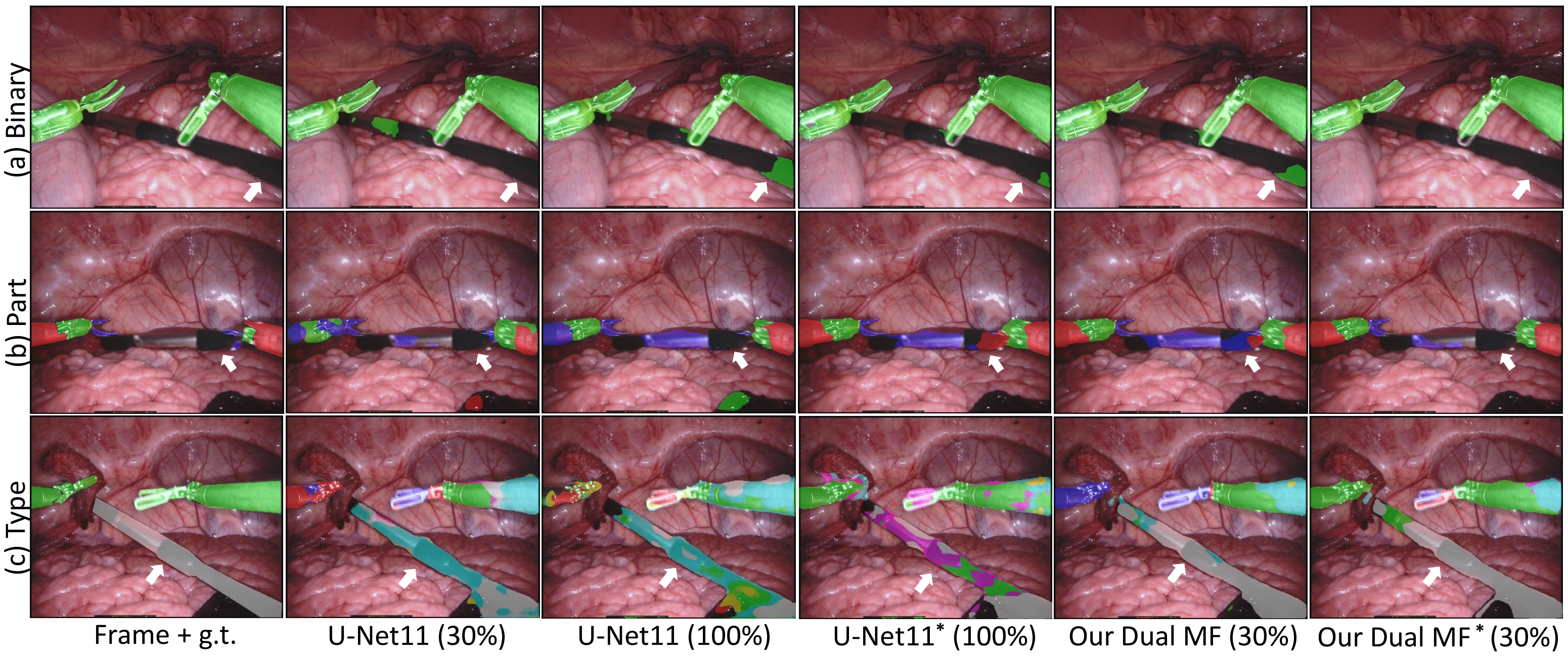}
	\caption{Visualization of instrument (a) binary, (b) part, and (c) type segmentation. From left to right, we present frame with ground-truth,  results of fully supervised training and our semi-supervised methods. $ ^{\star} $ denotes incorporating ConvLSTM units.}
	\label{fig:seg}
\end{figure}
\\
\textbf{Analysis of Our Semi-supervised Methods.}
For 6$ \times $225-frame training videos in each fold, we study the frames labeled at an interval of 2, 4, and 8, resulting in 30\%, 20\%, and 10\% annotations.
Table \ref{tab2} also lists results of three ablation settings: (1) Our Single MF: U-Net11 trained by set $ \{\mathcal{D}_{L}\!\cup\!\mathcal{D}_{R}\} $ with Flow Prediction; (2) Our Dual MF: U-Net11 trained by set $ \{\mathcal{D}_{L}\!\cup\!\mathcal{D}_{R}\!\cup\!\mathcal{D}_{C}\} $ with Flow Prediction and Compensation; 
(3) Our Dual MF$ ^{\star} $: U-Net11 embedded with ConvLSTM and trained by set $ \{\mathcal{D}_{L}\cup\mathcal{D}_{R}\cup\mathcal{D}_{C}\} $.   
It is observed that under all annotation ratios, compared with U-Net11 trained by labeled set $ \mathcal{D}_{L} $ alone, our flow based framework can progressively boost the semi-supervised performance with generated annotations.
We gain the maximum benefits in the severest condition (10\% labeling), where our Single MF has already largely improved the segmentation by 6.12\% IoU and 4.14\% Dice (binary), 2.67\% IoU and 1.50\% Dice (part), 14.37\% IoU and 20.82\% Dice (type).
Leveraging compensated pairs, our Dual MF with 30\% and 20\% labels is even able to exceed the full annotation training by 1-3\% IoU or Dice, corroborating that our method can recover and adjust the motion distribution  for better network training. 
It can be further verified using  temporal units. We only see slight improvements in fully supervised setting (the first two rows) because some motion inconsistency existed in original videos decreases the model learning capability of temporal cues.
Excitingly, the increment is obvious between our Dual MF and Dual MF$ ^{\star} $, especially for the toughest  type segmentation.
For instance, IoU and Dice can be boosted by 6.65\% and 6.25\% in 20\% labeling case. 
Fig.~\ref{fig:seg} shows some visual results. Our Dual MF$^{\star}$ can largely suppress misclassified regions \begin{figure}[!h]
	\centering
	\includegraphics[width=0.8\textwidth]{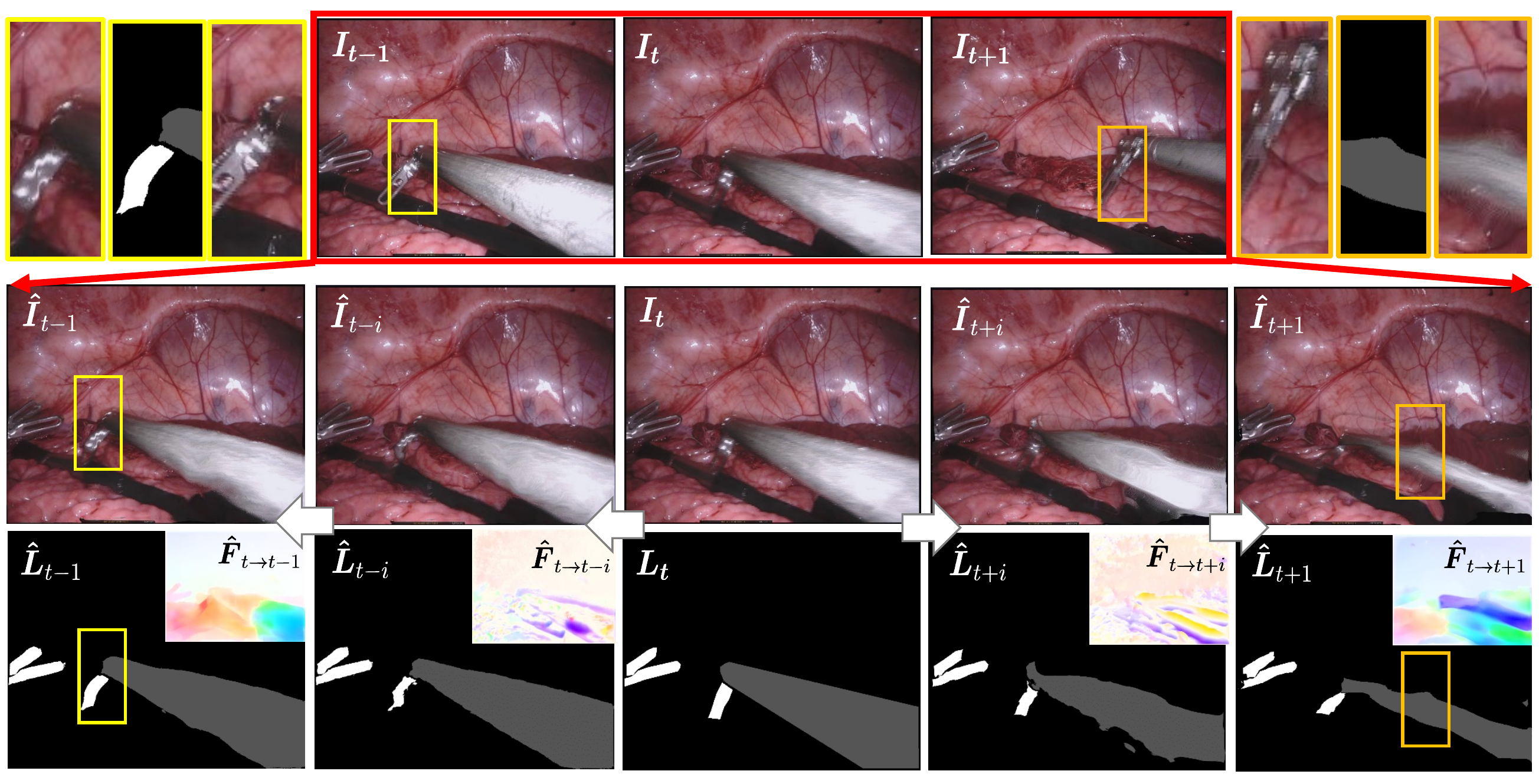}
	\caption{Example of rearranged training sequence with propagation step $k=1$.}
	\label{fig:gen}
\end{figure}
in Ultrasound probes for binary and part tasks, and achieve more complete and consistent type segmentation. 
It is even better at distinguishing hard mimics between instruments than fully supervised U-Net11$^{\star}$.
\\
\textbf{Analysis of Frame-Label Pairs.} 
Our joint propagation can alleviate the mis-alignment issue from label propagation. 
In Fig.~\ref{fig:gen}, labels in certain regions, like jaw (yellow) and shaft (orange) of instruments, fail to align with the original frames (first row) due to imprecision in learned flows, but correspond well with propagated frames (second row) as they experience the same transformation. The good alignment is crucial for segmentation.
Besides, our learned flows can propagate instruments to a more reasonable position with smooth motion shift. 
The fast instrument motion is slowed down from $ I_{t} $ to $ \hat{I}_{t+1} $ with smoother movement of Prograsp Forceps (orange), greatly benefiting ConvLSTM training.

\section{Conclusions}
We propose a flow prediction and compensation framework for semi-supervised instrument segmentation. Interestingly, we study the sparsely annotated  surgical  videos from the fresh perspective of learning the motion flow.
Large performance gain over state-of-the-art semi-supervised methods demonstrates the effectiveness of our framework. Inherently our method can recover the temporal structure of raw videos and be applied to surgical videos with high motion inconsistency.

\subsubsection{Acknowledgments.}
This work was supported by Key-Area Research and Development Program of Guangdong Province, China (2020B010165004),
Hong Kong RGC TRS Project No.T42-409/18-R, 
National Natural Science Foundation of China with Project No. U1813204, and  CUHK Shun Hing Institute of Advanced Engineering (project MMT-p5-20).

.

%
%
%
\bibliographystyle{splncs04}
\bibliography{refs}

\end{document}